\documentclass[sigconf]{acmart}
\AtBeginDocument{%
  }

\copyrightyear{2026}
\acmYear{2026}
\setcopyright{cc}
\setcctype{by}
\acmConference[MM '26]{Proceedings of the 34th ACM International Conference on Multimedia}{November 10--14, 2026}{Rio de Janeiro, Brazil}
\acmBooktitle{Proceedings of the 34th ACM International Conference on Multimedia (MM '26), November 10--14, 2026, Rio de Janeiro, Brazil}
\acmDOI{10.1145/3767308.3835990}
\acmISBN{979-8-4007-2213-4/2026/11}

\acmSubmissionID{mfp5977}

\usepackage{makecell}
\usepackage{algorithm}
\usepackage{algpseudocode}
\usepackage{amsmath} 
\usepackage{placeins}
\usepackage{multirow} 
\begin{document}

\title{CSGen: A Multi-Domain Curvilinear Structure Generation Model via Hierarchical Multimodal Diffusion}

\author{Zhe Shan}
\email{shanzard@hainanu.edu.cn}
\orcid{0009-0004-7979-8226}
\affiliation{%
  \institution{Hainan University}
  \city{Haikou}
  \country{China}}

\author{Ziming Yang}
\email{18925291090@stu.gdou.edu.cn}
\orcid{0009-0001-7014-7361}
\affiliation{%
  \institution{Guangdong Ocean University}
  \city{Zhanjiang}
  \country{China}}

\author{Lei Zhou}
\authornote{Lei Zhou and Xia Xie are corresponding authors.}
\email{leizhou@hainanu.edu.cn}
\orcid{0000-0003-0860-0724}
\affiliation{%
  \institution{Hainan University}
  \city{Haikou}
  \country{China}}

\author{Wenwen Zhang}
\email{zhangwenwen1000@gmail.com}
\orcid{0000-0002-3183-0344}
\affiliation{%
  \institution{Hainan University}
  \city{Haikou}
  \country{China}}

\author{Cong Lin}
\email{lincong@gdou.edu.cn}
\orcid{0000-0002-1567-1398}
\affiliation{%
  \institution{Guangdong Ocean University}
  \city{Zhanjiang}
  \country{China}}

\author{Xia Xie}
\authornotemark[1]
\email{shelicy@hainanu.edu.cn}
\orcid{0009-0002-6890-3663}
\affiliation{%
  \institution{Hainan University}
  \city{Haikou}
  \country{China}}

\renewcommand{\shortauthors}{Shan et al.}

\begin{abstract}
Curvilinear structure analysis is an important and fundamental task in multimedia. However, the controllable generation of images with precise curvilinear structure objects remains an open challenge. To address this, we propose CSGen, a hierarchical multimodal diffusion model that synthesizes high-fidelity images precisely aligned with multiple control conditions. The CSGen is built upon three key innovations: 1) We construct a multi-domain and multimodal dataset, including over 24K samples from 5 domains and 7 different types of annotations, to train the unified generation model. 2) We propose a novel hierarchical progressive control strategy that decouples topology clues from visual context by a phased signal injection, mitigating semantic drift while ensuring the topological integrity of sparse structures. 3) We design a sparsity-aware loss re-weighting mechanism to address the extreme sparsity of curvilinear structures, significantly enhancing the attention on thin and fragile structures during optimization. Extensive experiments demonstrate that CSGen generates images with superior structure accuracy and visual realism, significantly improving downstream segmentation performance while maintaining robustness across diverse prompts. Our results confirm CSGen as a scalable, data-centric paradigm for the analysis of complex curvilinear structures in diverse multimedia applications. Code and dataset are available at https://github.com/ShanZard/CSGen. 
\end{abstract}





\maketitle

\section{Introduction}
% Curvilinear structure analysis is an important and fundamental  task in multimedia, which is widely used from medical images to remote sensing. However, the controllable generation of images with precise curvilinear structure objects remains an unresolved problem. However, the extreme thinness, spatial sparsity, and complex topology of these structures pose significant challenges for existing algorithms. Furthermore, the high cost of pixel-level annotation often results in a scarcity of high-quality training data. 

\begin{figure}[t]
  \centering
  \includegraphics[width=0.85\linewidth]{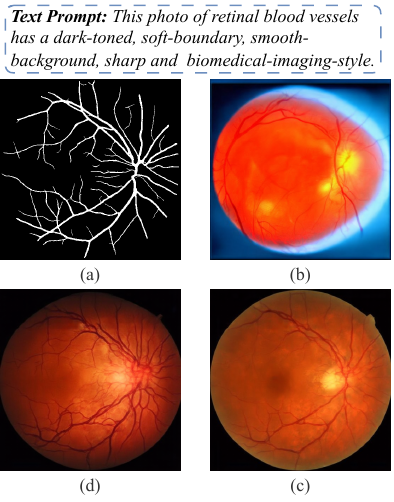}
  \Description{A four-panel comparison of curvilinear structure generation. Panel (a) shows the input layout map and text prompt. Panels (b) and (c) display results from SD3 with LoRA and ControlNet, where fine-grained vessels are fragmented. Panel (d) shows the CSGen result, which maintains perfect connectivity and alignment with the input.}
  \caption{Given the input text prompt and layout map (a), existing SOTA methods such as (b) SD3.5 with LoRA-based fine-tuning and (c) SD3.5 with ControlNet fail to reconstruct fragile, fine-grained branches, resulting in significant structure misalignment. Our CSGen (d) achieves strict topological connectivity and precise semantic adherence, even for the most sparse and delicate structures.}
  \label{fig:intro}
\end{figure}

Curvilinear structure analysis is a fundamental task in various multimedia processing and analysis~\cite{Cheng_2021_ICCV, 9843860, 10889719}. Precise curvilinear structure extraction is essential for several applications, spanning automated vascular analysis in medical imaging~\cite{LUO2025111254,10887048}, road network segmentation in remote sensing~\cite{ZHAO2024104082, LIU2024102009,11093073}, crack-based structural health monitoring~\cite{KABOODKHANI2025103271,mmxie2025}, etc~\cite{11243407,SHAN2025111162,LIAN2026131686}. Compared to the advancements in object segmentation, the controllable generation of high-quality images with complex curvilinear objects remains an open and under-explored problem in multimedia.

Unlike common objects, curvilinear structure objects are characterized by elongated geometric profiles and rigorous topological connectivity~\cite{Liu_Zhao_Zheng_2024}. These objects typically occupy a small fraction of the total image pixels, resulting in extreme spatial sparsity and a severe foreground-background signal imbalance. The distinctive characteristics pose huge challenges for deep learning models, which often struggle to maintain the continuity and precision required for fine-grained curvilinear signals. Furthermore, the high cost of pixel-level annotation of such special structures often results in a scarcity of high-quality training data.

Diffusion-based generative models have achieved great success in synthesizing objects with coarse-grained semantic signals, such as humans, vehicles, and natural landscapes~\cite{Rombach_2022_CVPR,10081412,10.1145/3626235}. However, when applied to curvilinear structure objects, these models encounter significant challenges, as their pre-trained priors fail to adapt to the spatial sparsity and intricate connectivity essential for high-fidelity generation. This raises the question: what distinguishes diffusion model performance in common objects from its limitations in curvilinear structure objects? We aim to address this question from three fundamental perspectives.

\textbf{(i) Data Scarcity and Annotation Bottlenecks:} The success of latent diffusion models in specialized domains relies on the high-quality dataset with precise semantic descriptors~\cite{NEURIPS2022_a1859deb,betker2023improving,10988859}. However, unlike natural scenes that possess rich object-level semantics and well-defined spatial hierarchies, curvilinear structures are often characterized by a monolithic composition, typically limited to a sparse foreground-background. Consequently, the absence of identifiable instances and structured relational dependencies makes it exceptionally difficult to construct coherent, high-quality image-text pairs.

\textbf{(ii) Semantic Drift in Fine-Grained Control:} The inherent morphological complexity of curvilinear structures defies conventional holistic text prompts. In multi-text encoder architectures such as Stable Diffusion 3.5 (SD3.5)~\cite{esser2024scaling}, employing uniform descriptors across disparate text encoders (e.g., CLIP~\cite{pmlr-v139-radford21a} and T5~\cite{JMLR:v21:20-074}) frequently precipitates semantic entanglement. Furthermore, a significant granularity dilemma exists in which simple prompts fail to provide sufficient geometric anchors, whereas overly verbose descriptions introduce linguistic noise that destabilizes the latent generation trajectory.

\textbf{(iii) Optimization Imbalance due to Spatial Sparsity:} The optimization of diffusion-based generation for curvilinear structures remains challenging because of the extreme foreground-background imbalance and the sparse, topology-sensitive nature of thin branches. Within the latent manifold of diffusion models, the standard MSE loss treats all spatial coordinates with uniform importance, which does not explicitly prioritize topology-critical regions~\cite{10.1007/978-3-031-72848-8_22,wang2023diffusion,falck2025fourier}. As a result, the model may under-emphasize thin, low-contrast structures and struggle to preserve connectivity in sparse or ambiguous regions.

In light of these analyses, we propose a hierarchical multimodal curvilinear structure generation model (CSGen) that achieves generating precise images from multiple types of prompts.  First, we establish a robust generation pipeline that integrates the SD3.5 backbone with structural control modules. For training this model, we construct a comprehensive dataset comprising over 24K samples across 5 domains, with 7 type annotations, providing a rich generative prior. Second, we propose a hierarchical progressive control strategy (HPCS) that discretizes continuous geometric attributes into stable semantic tokens to regularize the latent space. It follows a progressive training schedule that prioritizes semantics and structure through CLIP-based attributes, subsequently refining the generation with rich environmental context via T5 embeddings. Finally, we design a sparsity-aware loss re-weighting mechanism (SLRM) that quantifies structure difficulty based on the equivalent diameter. This mechanism improves the attention on thin, fragile, and topologically challenging structures during optimization.

The main contributions can be summarized as follows:

\begin{enumerate}
\item  We propose CSGen and construct a comprehensive dataset comprising 24K+ samples across 5 domains and 7 annotation types, providing a robust prior knowledge to address the data-starvation challenge in curvilinear structure analysis.

\item A hierarchical progressive control strategy is introduced that combines attribute discretization with progressive scheduling, effectively decoupling geometric topology from visual context to mitigate semantic drift and topological fragmentation.

\item We design a sparsity-aware re-weighting mechanism based on equivalent diameter, which encourages the model to focus on fragile, small-scale structures during training, significantly enhancing boundary precision and topological consistency.
\end{enumerate}

\section{Related Work}

\subsection{Curvilinear Structure Analysis}

Curvilinear structure analysis is a long-standing topic in multimedia, with representative applications spanning retinal vessels, coronary arteries, roads, and cracks~\cite{BIBILONI2016949}. Given the unique challenges of the curvilinear structure, many studies have explored ways to enhance its accuracy and robustness from multiple perspectives. Cheng et al.~\cite{Cheng_2021_ICCV} propose JTFN, which combines global topology modeling with feature refinement by iterative feedback. DSCNet~\cite{Qi_2023_ICCV} enhances curvilinear structure modeling using dynamic snake convolution, multi-view feature fusion, and a continuity-constraint loss to preserve topology. To reduce manual labeling, SemiCurv~\cite{9843860} employs a consistency-based semi-supervised framework with geometric transformations and N-pair loss, achieving strong performance with only 5\% labeled data. Recently, Mo et al.~\cite{10889719} propose semi-supervised curvilinear segmentation by leveraging distribution-alignment-informed thresholding to enhance the fidelity of pseudo-labels. Lei et al.~\cite{ECCVCCS} propose using diffusion models to generate data to improve downstream segmentation tasks, with the core idea of aligning semantic maps and images through textual descriptions of some curve object features. Although these methods have improved segmentation quality, most existing studies focus on enhancing discriminative prediction rather than controllable generative modeling. In particular, they do not address how to construct large-scale structured data priors or how to jointly regulate topology and appearance during generation, which are central to our setting.

\subsection{Diffusion Model} 

Diffusion models have become a dominant paradigm in modern generative modeling due to their strong synthesis quality, stable training, and flexibility in controlling conditions. Early formulations are developed from three closely related perspectives, including denoising diffusion probabilistic models (DDPMs)~\cite{NEURIPS2020_4c5bcfec}, score-based generative models (SGMs)~\cite{NEURIPS2019_3001ef25}, and stochastic differential equation-based models (Score-SDEs)~\cite{song2021scorebased}. Although these views differ in parameterization and theoretical interpretation, they are now widely understood as a unified family of iterative generative processes. 
The most relevant topic to our research is layout to image~\cite{Zhao_2019_CVPR,zhao2020layout2image,2023mmlayou,goal2024}. It aims to synthesize realistic images conditioned on structured spatial layouts. Zheng et al.~\cite{Zheng_2023_CVPR} introduce a dedicated layout fusion mechanism and object-aware cross attention to better model spatial relationships and object details, achieving superior control and fidelity. 
Guided Image Synthesis via Initial Image Editing in Diffusion Model further shows that manipulating the initial latent can provide effective spatial control for layout-aware synthesis~\cite{mao2023guided}.
Moving further, Xue et al.~\cite{Xue_2023_CVPR} further explore open-world generalization by injecting pre-trained text-to-image diffusion priors with a novel rectified cross-attention module. Shabani et al.~\cite{Shabani_2024_CVPR} introduce an image-vector dual diffusion framework that operates jointly in image and vector spaces, enhancing inter-element reasoning and visual consistency. The existing methods are primarily developed for object-centric scenes or design elements with relatively dense visual support. They are not tailored to curvilinear structures, whose generation is dominated by extreme sparsity, fragile connectivity, and strong coupling between geometry and appearance. Our work addresses this gap by developing a diffusion-based model specialized for controllable curvilinear structure generation.

% Subsequent research has mainly advanced diffusion models along three directions: (1) accelerating and improving the efficiency of sampling procedures~\cite{NEURIPS2022_a98846e9,Meng_2023_CVPR}, (2) achieving more accurate likelihood and density estimation~\cite{pmlr-v162-lu22f,huang2024proteinligand}, and (3) extending diffusion models to data domains with stronger structural constraints, such as permutation invariance, manifold constraints, or discrete representations~\cite{NEURIPS2022_df04a35d,Gu_2022_CVPR,Yang_2024_CVPR}. 

\section{CSGen}\label{sec:format}

%% 多领域数据

\subsection{Preliminary: Diffusion Model and Stable Diffusion}

We build CSGen on Stable Diffusion 3.5, which follows the SD3 rectified-flow transformer design~\cite{esser2024scaling}. The model operates in the latent space of a pre-trained VAE: the encoder maps an image $x_0$ to a latent $z_0$, and the decoder maps the generated latent back to RGB space. Text conditions $\mathcal{C}$ are encoded by multiple pre-trained text encoders, while the MM-DiT backbone models joint interactions between image and text tokens.

Unlike conventional DDPMs that predict Gaussian noise, SD3 learns a velocity field along a straight interpolation path between data latents and Gaussian noise latents. Given $z_0$, a noise latent $z_1 \sim \mathcal{N}(0, I)$, and $t \in [0,1]$, we use
\begin{equation}
    z_t = (1-t)z_0 + t z_1, \quad v = z_1 - z_0 .
\end{equation}
The flow matching objective is formulated as
\begin{equation}
    \mathcal{L}_{FM} =
    \mathbb{E}_{z_0,z_1,t,\mathcal{C}}
    \left[
    \left\|
    v - v_{\theta}(z_t,t,\mathcal{C})
    \right\|_2^2
    \right],
    \label{eq:fm_objective}
\end{equation}
where $v_{\theta}$ denotes the velocity predicted by the MM-DiT denoiser. During sampling, the model starts from a Gaussian latent and integrates the learned velocity field backward toward the data latent under condition $\mathcal{C}$ before VAE decoding.

\begin{figure*}[t]
  \centering
  \includegraphics[width=0.95\linewidth]{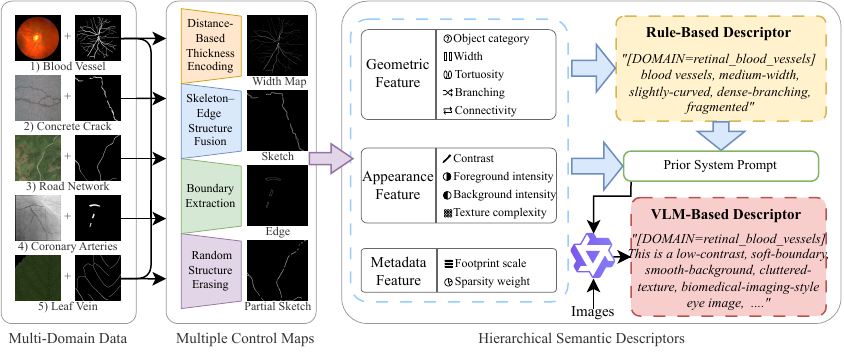}
  \Description{A flowchart illustrating the dataset construction pipeline. It shows raw images and masks being converted into multiple control maps, which are then analyzed to create hierarchical rule-based and VLM-based text descriptors for the final generation prompt.}
  \caption{Illustration of constructing the dataset. Multi-domain curvilinear images and masks are transformed into multiple control maps. They are further summarized as hierarchical semantic descriptors through geometric, appearance, and metadata features. The resulting rule-based and VLM-based descriptors are finally integrated as the generation prompt.}

  \label{fig:fig1}
\end{figure*}
\subsection{Overview of the CSGen}

As shown in Figure~\ref{fig:fig1}, we first build different types of annotations from the original dataset, including structural cues that describe geometric and topological information and text descriptions that provide semantic and appearance information. Then, these prompts are decoupledly fed into the multi-text encoder and VAE, only training the ControlNet branch, while the frozen SD3.5 backbone and text encoder provide pretrained generative knowledge, as shown in Figure~\ref{fig:fig2}. During training, we use the hierarchical progressive control strategy to gradually introduce different types of prompts, helping the model learn the relationship between language and structure. Meanwhile, we introduce a sparsity-aware loss re-weighting mechanism. This allows the model to focus more on difficult samples and improves both structural consistency and generation quality. Finally, the trained CSGen can accept prompts of any style and type to generate high-quality images for downstream tasks.

\subsection{Preparation Dataset}
\label{sec:dataset}

The quality and diversity of the training data serve as the primary determinants of the model's generative performance. However, large-scale image-text datasets used for diffusion pre-training are dominantly biased toward natural objects and scenes, prioritizing holistic appearance over fine-grained structural precision. Based on this, we build a high-quality dataset spanning 5 representative domains of curvilinear structures: retinal blood vessels, coronary arteries, concrete cracks, leaf veins, and road networks, as shown in Table~\ref{tab:dataset_statistics}. Each sample comprises the original image and its corresponding masks. For each image, we generate various structure control maps from masks in different ways. Meanwhile, textual prompts and morphological structure information are also constructed for each sample. The integration of these domains yields a twofold benefit. It substantially augments the training diversity while capitalizing on the shared morphological characteristics inherent in all curvilinear structures.

\begin{table}[!t]
\centering
\caption{Statistics of the collected multi-domain curvilinear structure dataset.}
\label{tab:dataset_statistics}
\footnotesize
\setlength{\tabcolsep}{3pt}
\renewcommand{\arraystretch}{1.05}
\begin{tabular}{p{2.2cm} p{4.0cm} c c}
\toprule
Domain & Source datasets & Count & Prop. (\%) \\
\midrule
Retinal Blood Vessel &
HRF~\cite{budai2013robust}, CHASE\_DB1~\cite{fraz2012ensemble}, STARE~\cite{hoover2000locating}, \newline
FIVES~\cite{jin2022fives}, RETA~\cite{lyu2022reta}, LES-AV~\cite{orlando2018towards}, DRIVE~\cite{staal2004ridge}
& 920 & 3.73 \\

Coronary Artery &
ARCADE~\cite{popov2024arcade}
& 2997 & 12.14 \\

Concrete Crack &
Concrete Crack Conglomerate~\cite{bianchi2021concrete}
& 8370 & 33.92 \\

Leaf Vein &
HALVS~\cite{LIU2026100164}
& 6165 & 24.98 \\

Road Network &
DeepGlobe~\cite{demir2018deepglobe}
& 6226 & 25.23 \\
\midrule
Total & -- & 24678 & 100.00 \\
\bottomrule
\end{tabular}
\end{table}

\subsubsection{Construction Multiple Structural Control Maps}

Curvilinear structures have two key characteristics. Their local boundaries must be clear, and the global structure must remain continuous and correctly oriented. To ensure local boundary precision and global topological continuity, we derive complementary structural supervision signals from each mask. As illustrated in Figure~\ref{fig:fig1}, we extract four specialized control maps. (1) Width Maps: Generated via distance-based thickness encoding to represent the variable caliber of the structures. (2) Sketches: Produced through skeleton-edge structure fusion to capture the unified geometric layout. (3) Edge Maps: Extracted using Canny edge detection to emphasize local contour precision and boundary localization. (4) Partial Sketches: Created via random structure erasing to improve the model's robustness against incomplete or occluded structural inputs. Significantly, this multiple control map strategy offers practical advantages beyond training stability and enables users to guide the generation through various intuitive modalities.

\subsubsection{Extraction Attribute and Meta-information}

To describe the morphology of curvilinear structures and their visual environments, we extract three distinct categories of features: geometric, appearance, and metadata features from images and masks.  As illustrated in Figure~\ref{fig:fig1}, these quantified attributes serve a dual purpose: they are formatted as explicit semantic descriptors to anchor the text-to-image mapping and utilized as structured prior system-level prompts.  By injecting these explicit structure priors into the training stage, we provide the model with stable geometric guidance that effectively mitigates the semantic ambiguity inherent in standard natural language prompts

\subsubsection{Generation Text Prompt}
High-quality image-text pairs are critical to ensure that the model correctly maps linguistic concepts to intricate geometric primitives. However, unlike natural objects that have clearly defined semantic information, it is unclear what aspects need to be described for curvilinear objects. To make the text prompts more stable, reasonable, and better suited for diffusion model training, we propose two specific levels of text prompts. First, we discretize key features within the same domain. Specifically, continuous features are divided into several discrete semantic labels. For example, we divided objects as straight, slightly curved, and tortuous based on tortuosity. Second, we employ a large vision-language model (VLM)~\cite{bai2025qwen3} to generate detailed environmental descriptions. Importantly, to prevent the VLM from generating hallucinated or structurally inconsistent text, we inject the precisely quantified metadata into the VLM as a prior system prompt. This constraint-aware prompting ensures that the VLM-generated captions remain synchronized with the underlying structural truth.

To ensure explicit distinction across these diverse domains, we employ a standardized prefixing convention for each prompt to explicitly denote the domain. Comprehensive details regarding the data construction pipeline and annotation protocols are provided in the supplementary materials.

\begin{figure*}[t]
  \centering
 \includegraphics[width=0.9\linewidth]{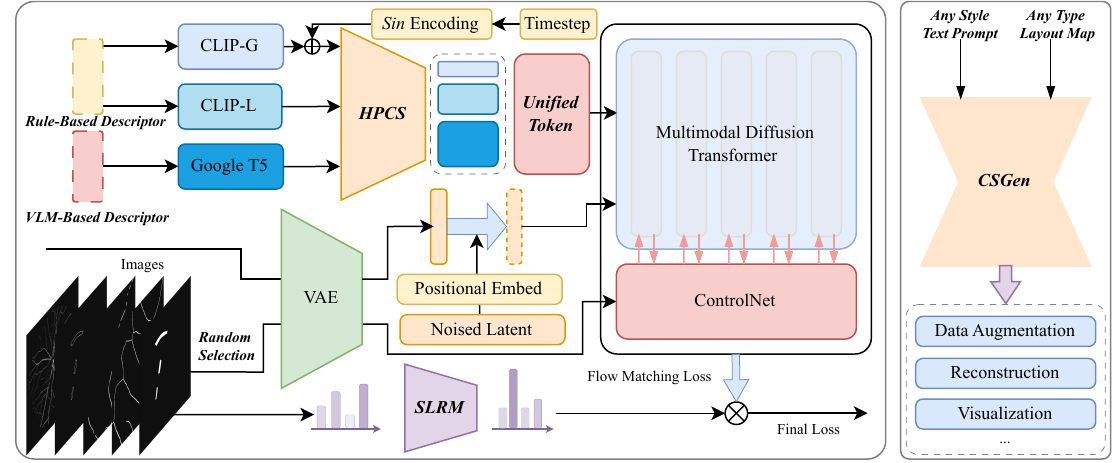}
 \Description{A system architecture diagram of CSGen. The left section shows the training pipeline involving VAE, HPCS for unified tokens, and SLRM weighting in a Diffusion Transformer. The right section shows inference and downstream applications like data augmentation and reconstruction.}
  \caption{Illustration of the CSGen. The training stage is illustrated on the left, while the inference stage and various downstream applications are shown on the right.}

  \label{fig:fig2}
\end{figure*}
\subsection{Training Diffusion Model}
\label{sec:training}

CSGen adopts the SD3.5 rectified-flow training process to curvilinear structure generation by adding structural conditioning and sparsity-aware re-weighting. We use the SD3.5 architecture as our generative backbone, integrated with a ControlNet to randomly inject structural maps $\mathcal{C}_{map} \in \{ \text{Mask, Width, Sketch, Edge, Partial} \}$. ControlNet augments a large pretrained diffusion model with spatially localized image conditions, enabling more controllable and structure-aware generation.

\subsubsection{Hierarchical Progressive Control Strategy}

In generative model training, the precision of text guidance strongly affects how the model aligns structure and appearance. We observed that attempting to synthesize intricate curvilinear structures and complex environmental textures simultaneously can make optimization unstable. Without a stable reference, the model struggles to distinguish between the core topological signal and the surrounding visual noise. To resolve this, we propose the hierarchical progressive control strategy (HPCS), as shown in Figure~\ref{fig:fig2}. 

We first establish a geometric anchor $\mathbf{h}_{geo}$ to stabilize the model's perception of domain-specific structure and semantics before introducing contextual refinements $\mathbf{h}_{ctx}$. Based on the decoupled architectural strengths of text encoders, we define:
\begin{equation}
\mathbf{h}_{geo} = \text{CLIP}(\mathcal{D}_{rule}), \quad \mathbf{h}_{ctx} = \text{T5}(\mathcal{D}_{vlm}),
\end{equation}
where $\mathbf{h}_{geo}$ captures stable geometric attributes and $\mathbf{h}_{ctx}$ encodes fine-grained scene details. To avoid linguistic noise in early training, the unified embedding $\mathbf{E}_{unified}(\tau)$ is formulated via a phased injection:
\begin{equation}
\mathbf{E}_{unified}(\tau) = \Phi(\mathbf{h}_{geo} ) + \lambda(\tau) \cdot \Psi(\mathbf{h}_{ctx}),
\end{equation}
where $\Phi$ and $\Psi$ are learned linear projections that align the CLIP and T5 features to the conditioning space of the diffusion transformer. The progressive weight $\lambda(\tau)$ is governed by a curriculum schedule: $\lambda(\tau) = \text{clip} ( \frac{\tau - \tau_{anchor}}{\tau_{total} \cdot \eta}, 0, 1 )^k$, where $\tau_{anchor}$ defines a topology-anchoring phase. This hierarchical decoupling helps the model first learn stable structural semantics before incorporating fine-grained environmental textures.

\begin{algorithm}[!t]
\caption{CSGen Hierarchical Multimodal Training}
\label{alg:csgen_training}
\begin{algorithmic}[1]
\Require 
    Multi-domain dataset $\mathcal{D} = \{(x_i, \mathcal{C}_i, \mathcal{D}_{rule}, \mathcal{D}_{vlm})\}_{i=1}^N$; Hyperparameters: $\tau_{total}, \tau_{anchor}, \eta, k, \gamma, \epsilon_0, D_{max}$.
\Ensure 
    Optimized parameters $\theta$ (ControlNet).

\State Initialize $\theta$ and freeze the weights of SD3.5, VAE, and text encoders.
\For{iteration $\tau = 1$ \textbf{to} $\tau_{total}$}
    \State \textbf{// 1. Rectified-flow latent interpolation}
    \State Sample batch $(x, \mathcal{C}_{map}, \mathcal{D}_{rule}, \mathcal{D}_{vlm}) \sim \mathcal{D}$;
    \State $z_0 \leftarrow \text{VAE.encode}(x)$;
    \State Sample $t \sim \mathcal{U}(0, 1)$ and noise latent $z_1 \sim \mathcal{N}(0, \mathbf{I})$;
    \State $z_t \leftarrow (1-t)z_0 + t z_1$;
    \State $v \leftarrow z_1 - z_0$;
    
    \State \textbf{// 2. Hierarchical Progressive Control Strategy}
    \State $\mathbf{h}_{geo} \leftarrow \text{CLIP}(\mathcal{D}_{rule})$; 
    \State $\mathbf{h}_{ctx} \leftarrow \text{T5}(\mathcal{D}_{vlm})$;
    \State Compute injection weight: $\lambda(\tau) = \text{clip} \left( \frac{\tau - \tau_{anchor}}{\tau_{total} \cdot \eta}, 0, 1 \right)^k$;
    \State $\mathbf{E}_{unified}(\tau) = \Phi(\mathbf{h}_{geo}) + \lambda(\tau) \cdot \Psi(\mathbf{h}_{ctx})$;
    
    \State \textbf{// 3. Sparsity-Aware Loss Re-weighting Mechanism}
    \State Calculate equivalent diameter $d_i = 2\sqrt{A_i / \pi}$ from structure layout $\mathcal{C}_{map}$;
    \State $\mathcal{W}(d_i) = \gamma \cdot \log \left( 1 + \frac{D_{max}}{d_i + \epsilon_0} \right)$;
    
    \State \textbf{// 4. Flow matching optimization}
    \State Predict velocity: $\hat{v} = v_{\theta}(z_t, t, \mathbf{E}_{unified}(\tau), \mathcal{C}_{map})$;
    \State $\mathcal{L}_{final} = \frac{1}{B} \sum_{i=1}^{B} \mathcal{W}(d_i) \cdot \left\| v_i - \hat{v}_i \right\|_2^2$;
    \State Update $\theta \leftarrow \theta - \nabla_{\theta} \mathcal{L}_{final}$ via AdamW;
\EndFor
\State \Return $\theta$.
\end{algorithmic}
\end{algorithm}

\begin{table*}[t]
\centering
\caption{Comparison with SOTA generative model under the same prompt setting.}
\label{tab:comparison}
\begin{tabular}{lcccccc}
\toprule
Method & FID $\downarrow$ & IS $\uparrow$ & LPIPS $\downarrow$ & DISTS $\downarrow$ & CLIPScore $\uparrow$ & VQA $\uparrow$ \\
\midrule
FLUX              & 204.357 $\pm$ 4.116 & \textbf{6.102 $\pm$ 0.349} & 0.847 $\pm$ 0.101 & 0.467 $\pm$ 0.064 & 0.256 $\pm$ 0.024 & 0.714 $\pm$ 0.452 \\
Qwen              & 186.432 $\pm$ 3.957 & 5.781 $\pm$ 0.428 & 0.889 $\pm$ 0.127 & 0.489 $\pm$ 0.061 & 0.254 $\pm$ 0.025 & 0.701 $\pm$ 0.458 \\
SD3.5               & 170.755 $\pm$ 3.814 & 5.946 $\pm$ 0.500 & 0.906 $\pm$ 0.135 & 0.498 $\pm$ 0.059 & 0.228 $\pm$ 0.032  & 0.680 $\pm$ 0.466 \\
SD3.5 + LoRA        & 124.786 $\pm$ 3.337 & 4.634 $\pm$ 0.295 & 0.671 $\pm$ 0.143 & 0.377 $\pm$ 0.068 & 0.245 $\pm$ 0.027 &  0.895 $\pm$ 0.307 \\
SD3.5 + ControlNet  & 131.256 $\pm$ 1.908 & 3.486 $\pm$ 0.169 & 0.667 $\pm$ 0.222 & 0.382 $\pm$ 0.076 & 0.224 $\pm$ 0.034 & 0.822 $\pm$ 0.375 \\
Ours              & \textbf{98.525 $\pm$ 2.688} & 4.188 $\pm$ 0.213 & \textbf{0.655 $\pm$ 0.211} & \textbf{0.375 $\pm$ 0.075} &  \textbf{0.258 $\pm$ 0.024}  & \textbf{0.960 $\pm$ 0.196}\\
\bottomrule
\end{tabular}
\end{table*}

\subsubsection{Sparsity-Aware Loss Re-weighting Mechanism}
Due to the extreme spatial sparsity of curvilinear structures, the structural foreground signals are frequently drowned out by dominant background noise gradients in standard loss functions. To address this signal imbalance, we propose the sparsity-aware loss re-weighting mechanism (SLRM) to improve the optimization focus. 

Specifically, we quantify the structural difficulty of each sample via the equivalent diameter $d_i = 2\sqrt{A_i / \pi}$, where $A_i$ represents the foreground area of the $i$-th mask. We define a sparsity-aware weight $\mathcal{W}(d_i)$ as:
\begin{equation}
\mathcal{W}(d_i) = \gamma \cdot \log \left( 1 + \frac{D_{max}}{d_i + \epsilon_0} \right),
\end{equation}
where $\gamma$ and $\epsilon_0$ are hyperparameters for scale control and numerical stability. Let $\hat{v}_i = v_{\theta}(z_{t,i}, t, \mathbf{E}_{unified}(\tau), \mathcal{C}_{map,i})$ denote the predicted velocity for the $i$-th sample. By incorporating this weight into the flow-matching objective in Eq.~\eqref{eq:fm_objective}, the final loss is formulated as:
\begin{equation}
\mathcal{L}_{final} =
\mathbb{E}_{z_0,z_1,t,\mathcal{C}_{map}}
\left[
\mathcal{W}(d_i) \cdot
\left\|
v_i - \hat{v}_i
\right\|_2^2
\right],
\end{equation}
where $v_i = z_{1,i}-z_{0,i}$. This mechanism assigns higher optimization weights to samples with thinner and more fragile structures (smaller $d_i$), encouraging the model to better preserve fine-grained topology under sparse foreground supervision. The complete optimization workflow is formally outlined in Algorithm~\ref{alg:csgen_training}.

\section{Experiments}
\label{sec:experiments}

\subsection{Implementation and Evaluation}
\label{sec:implementation}

We evaluate the generated results using two groups of metrics. The first group contains six image-level metrics: Fr\'echet Inception Distance (FID)~\cite{heusel2017ttur}, Inception Score (IS)~\cite{salimans2016improved}, LPIPS~\cite{zhang2018lpips}, DISTS~\cite{ding2022dists}, CLIPScore~\cite{hessel2021clipscore}, and a VQA-based score~\cite{lin2024vqascore}. FID and IS are used to measure overall generation quality and diversity. LPIPS and DISTS are used to measure perceptual similarity between generated images and reference images. CLIPScore and the VQA-based score are used to evaluate image-text semantic alignment.

However, these image-level metrics are still not enough for sparse curvilinear structure generation, because they cannot fully reflect local structure properties such as connectivity, breakage, and topology preservation. Therefore, we further use a parser-based structural evaluation. Specifically, we apply SegFormer~\cite{xie2021segformer} to segment the generated images and compare the parsed masks with the corresponding reference masks. Based on the parsed results, we report mIoU and mDice as region-overlap metrics, and tprec, tsens, and clDice as topology-aware metrics~\cite{shit2021cldice}.

We conduct experiments on generation models and discriminative models in the diffusers~\cite{von-platen-etal-2022-diffusers} and mmsegmentation~\cite{mmseg2020} code repositories, respectively. We follow the default configurations of these two libraries for most loss functions, optimizers, and hyperparameters. For further details, please refer to the supplementary material.

\subsection{Comparison Experiment}
\label{sec:comparison}

\begin{figure*}[!t]
  \centering
  \includegraphics[width=0.9\linewidth]{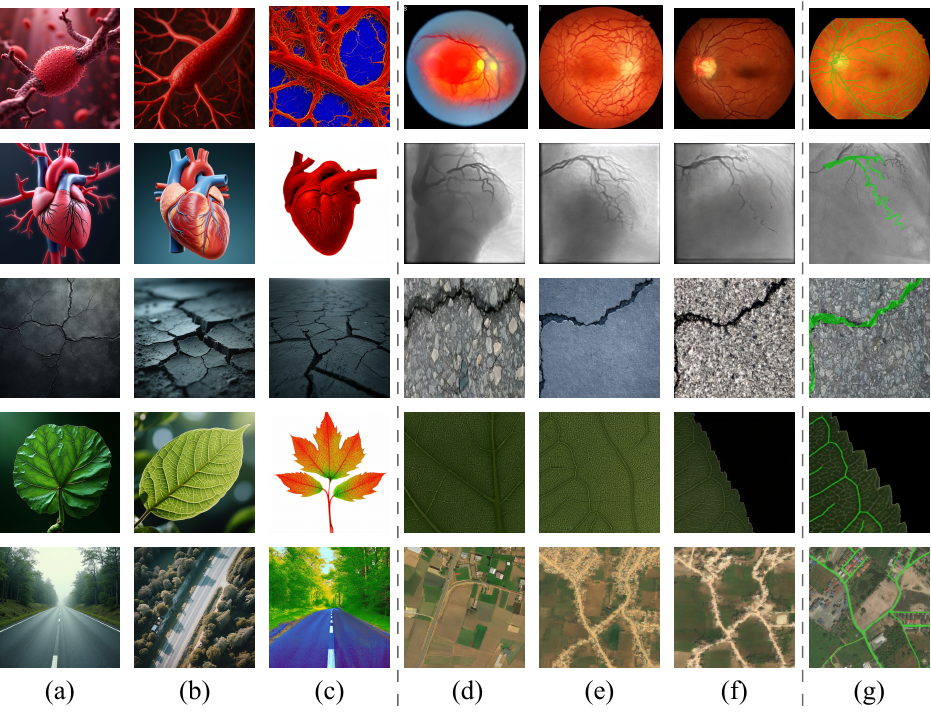}
  \Description{A grid showing qualitative results for various domains. Rows represent different categories like retinal vessels and roads. Columns compare training-free models, fine-tuned baselines, and our CSGen. CSGen produces the cleanest and most complete structures compared to the fragmented outputs of other models.}
  \caption{Qualitative results of curvilinear structure generation. (a-c) Training-free general models (FLUX, Qwen, SD3.5); (d-e) fine-tuned variants (SD3.5+LoRA, SD3.5+ControlNet); (f) our CSGen; and (g) raw images and corresponding layout maps. Our method shows significantly cleaner and more complete structure synthesis.}
  \label{fig:show}
\end{figure*}

Table~\ref{tab:comparison} compares our method with several representative generation models under the same prompt setting. Overall, our approach achieves SOTA performance across multiple metrics. We observe that general models without domain-specific training yield poor results, whereas fine-tuning significantly enhances image fidelity. Notably, FLUX exhibits the highest diversity, likely stemming from its diverse pre-training data distribution. However, while fine-tuning improves fidelity, it tends to incur a slight degradation in generative diversity. Figure~\ref{fig:show} further illustrates that our method produces cleaner and more structurally intact curvilinear patterns. Specifically, the intricate segments exhibit stringent spatial adherence to the input masks, effectively preserving topological connectivity even at a granular level.

\begin{table}[!t]
\centering
\caption{Topology-aware ablation under the fixed Mask control condition.}
\label{tab:topology_ablation}
% \scriptsize
% \setlength{\tabcolsep}{3pt}
% \renewcommand{\arraystretch}{1.02}
\begin{tabular}{lcccc}
\toprule
Method & mIoU$\uparrow$ & tprec$\uparrow$ & tsens$\uparrow$ & clDice$\uparrow$ \\
\midrule
Baseline & 0.431 & 0.791 & 0.578 & 0.628 \\
+ SLRM only & 0.462 & 0.797 & 0.650 & 0.659 \\
+ Text Decoupling & 0.439 & 0.793 & 0.605 & 0.645 \\
+ Text Decoupling + Prog. & 0.462 & 0.805 & 0.689 & 0.702 \\
Full Method & \textbf{0.487} & \textbf{0.821} & \textbf{0.757} & \textbf{0.766} \\
\bottomrule
\end{tabular}
\end{table}

\subsection{Ablation Analysis}
\label{sec:ablation}

\begin{table*}[!t]
\centering
\caption{Progressive ablation of the core components under a fixed Mask control condition.}
\label{tab:core_ablation}
\begin{tabular}{lcccccc}
\toprule
Method & FID$\downarrow$ & IS$\uparrow$ & LPIPS$\downarrow$ & DISTS$\downarrow$ & CLIPScore$\uparrow$ & VQA$\uparrow$ \\
\midrule
Baseline                         & 131.256 $\pm$ 1.908 & 3.486 $\pm$ 0.169 & 0.667 $\pm$ 0.222 & 0.382 $\pm$ 0.076 & 0.224 $\pm$ 0.034 & 0.822 $\pm$ 0.375 \\
+  Hierarchical Control Strategy          & 127.939 $\pm$ 4.212 & 3.672 $\pm$ 0.209 & 0.662 $\pm$ 0.218 & 0.380 $\pm$ 0.074 & 0.225 $\pm$ 0.033 & 0.835 $\pm$ 0.354 \\
+ Progressive Training       & 112.234 $\pm$ 3.689 & 4.143 $\pm$ 0.204 & 0.658 $\pm$ 0.214 & 0.377 $\pm$ 0.072 & 0.226 $\pm$ 0.033 & 0.874 $\pm$ 0.318 \\
+ Sparse Re-weighting  & \textbf{104.786 $\pm$ 3.337} & \textbf{4.188 $\pm$ 0.213} & \textbf{0.655 $\pm$ 0.211} & \textbf{0.375 $\pm$ 0.075} & \textbf{0.228 $\pm$ 0.032} & \textbf{0.895 $\pm$ 0.307} \\
\bottomrule
\end{tabular}
\end{table*}

The efficacy of each module is progressively validated in Table~\ref{tab:topology_ablation} and Table~\ref{tab:core_ablation}. Starting from a vanilla SD3.5-ControlNet baseline, we observe that hierarchical text decoupling provides better structural grounding, leading to improved fidelity. The introduction of progressive training further stabilizes the flow matching optimization by transitioning from coarse structural anchors to richer contextual details. Most notably, the introduction of SLRM provides the most significant performance boost, particularly in FID and VQA, showing that re-weighting the loss to focus on sparse, thin branches is critical for preserving topological integrity. The SLRM-only variant improves clDice from 0.628 to 0.659, while the full model reaches 0.766, indicating that the proposed re-weighting and hierarchical conditioning contribute to thin-structure continuity beyond image-level fidelity metrics.

% \begin{table*}[!t]
% \centering
% \caption{Ablation results under different structural control signals. }
% \label{tab:ablation_main}
% \begin{tabular}{lcccccc}
% \toprule
% Method & FID$\downarrow$ & IS$\uparrow$ & LPIPS$\downarrow$ & DISTS$\downarrow$ & CLIPScore$\uparrow$ & VQA$\uparrow$ \\
% \midrule
% Mask & 104.269 $\pm$ 3.716 & 4.171 $\pm$ 0.156 & 0.654 $\pm$ 0.210 & 0.374 $\pm$ 0.076 & 0.227 $\pm$ 0.031 & 0.894 $\pm$ 0.308 \\
% Edge & 104.950 $\pm$ 3.596 & 4.178 $\pm$ 0.280 & 0.656 $\pm$ 0.209 & 0.377 $\pm$ 0.077 & 0.228 $\pm$ 0.032 & 0.891 $\pm$ 0.312 \\
% Partial Sketch  & 104.360 $\pm$ 3.152 & 4.074 $\pm$ 0.146 & 0.658 $\pm$ 0.210 & 0.378 $\pm$ 0.076 & 0.228 $\pm$ 0.032 & 0.884 $\pm$ 0.320 \\
% Sketch  & 105.066 $\pm$ 3.419 & \textbf{4.211 $\pm$ 0.216} & \textbf{0.653 $\pm$ 0.210}& 0.374 $\pm$ 0.076 & 0.228 $\pm$ 0.032 & 0.892 $\pm$ 0.310 \\
% Width Map   & \textbf{103.806 $\pm$ 3.627} & 4.108 $\pm$ 0.188 & 0.659 $\pm$ 0.212 & 0.378 $\pm$ 0.076 & 0.227 $\pm$ 0.032 & 0.893 $\pm$ 0.309 \\
% Random  & 104.786 $\pm$ 3.337 & 4.188 $\pm$ 0.213 & 0.655 $\pm$ 0.211 & \textbf{0.373 $\pm$ 0.075} & \textbf{0.228 $\pm$ 0.033} & \textbf{0.895 $\pm$ 0.307} \\
% \bottomrule
% \end{tabular}
% \end{table*}

\begin{figure}[t]
  \centering
  \includegraphics[width=\linewidth]{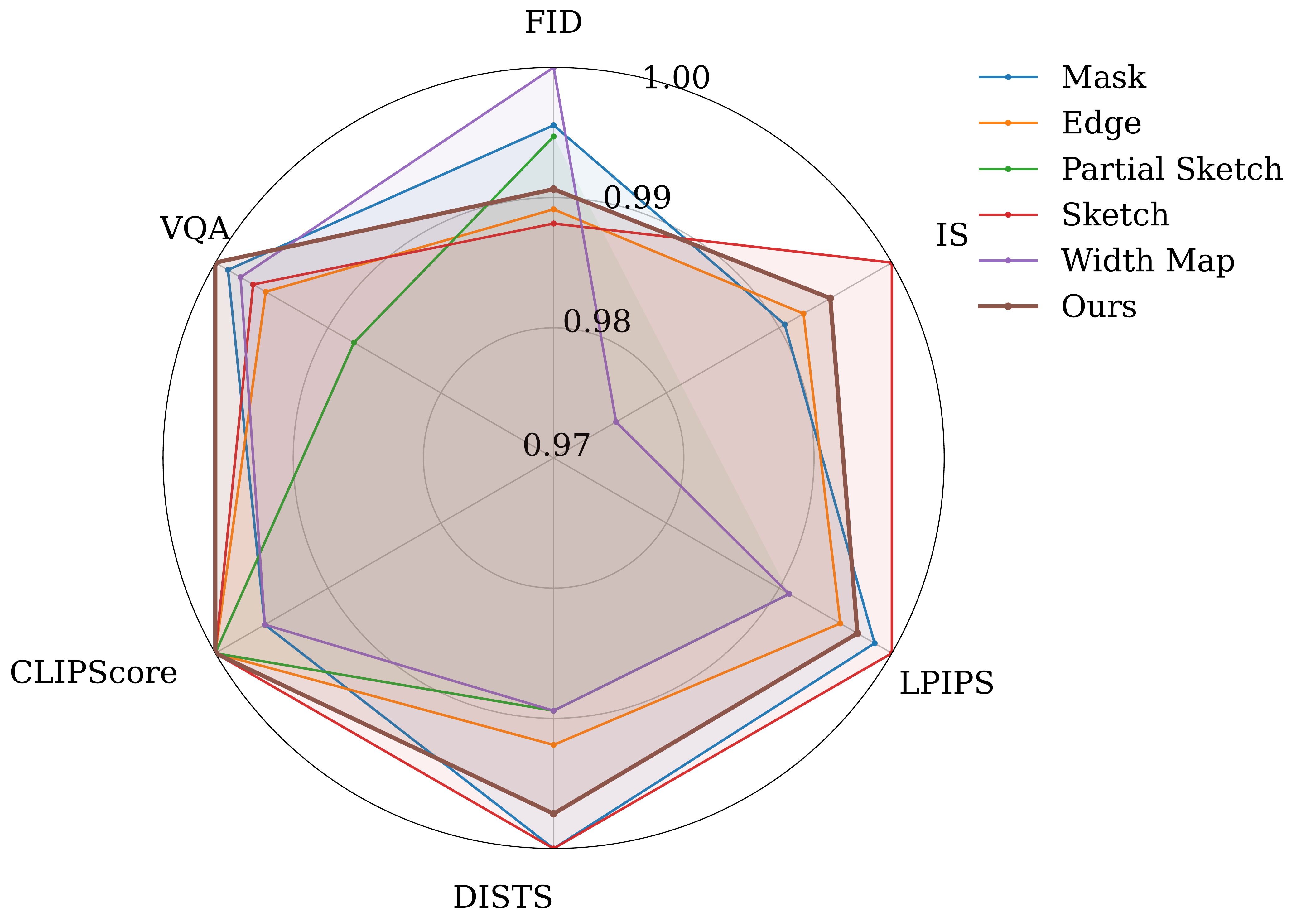}
\Description{A per-metric normalized radar chart showing the performance of different structural control conditions. The colored profiles for various settings largely overlap, indicating that the CSGen framework maintains stable performance across diverse structural priors.}
  \caption{Per-metric normalized radar visualization of different structural control conditions. Each metric is independently normalized across all compared methods. }
  \label{fig:control_signal}
  \vspace{-10pt}
\end{figure}

After the progressive ablation study, we further test whether the model remains stable under different control inputs. Since curvilinear generation may rely on different types of geometric priors, it is important to verify that the framework does not depend too much on one specific control condition. The results in Figure~\ref{fig:control_signal} show that the framework remains generally stable across different structural control settings. Interestingly, we observe that the IS exhibits a higher degree of susceptibility to changes in control signals compared to other metrics, indicating that the choice of structural prior significantly influences the semantic richness of the output. Overall, these results indicate that the proposed model can work well with multiple types of structural priors rather than relying on only one specific input condition.

\subsection{Curvilinear Structure Validity after Parsing}
\label{sec:parser_structure}

\begin{table*}[!t]
\centering
\caption{Parser-based structure evaluation under different control settings.}
\label{tab:structure_ablation}
\begin{tabular*}{\textwidth}{@{\extracolsep{\fill}}lccccc}
\toprule
Method & mIoU$\uparrow$ & mDice$\uparrow$ & tprec$\uparrow$ & tsens$\uparrow$ & clDice$\uparrow$ \\
\midrule
\multicolumn{6}{l}{\textit{Single structure control signals}} \\
\midrule
Edge            & 0.433 $\pm$ 0.214 & 0.569 $\pm$ 0.240 & 0.789 $\pm$ 0.188 & 0.640 $\pm$ 0.318 & 0.657 $\pm$ 0.288 \\
Partial Sketch  & 0.435 $\pm$ 0.192 & 0.579 $\pm$ 0.211 & 0.797 $\pm$ 0.174 & 0.647 $\pm$ 0.280 & 0.669 $\pm$ 0.244 \\
Sketch          & 0.479 $\pm$ 0.198 & 0.620 $\pm$ 0.209 & 0.823 $\pm$ 0.171 & 0.705 $\pm$ 0.276 & 0.720 $\pm$ 0.240 \\
Width Map       & 0.498 $\pm$ 0.196& 0.639 $\pm$ 0.199& 0.827 $\pm$ 0.179 & 0.711 $\pm$ 0.263& 0.728 $\pm$ 0.233 \\
\midrule

\multicolumn{6}{l}{\textit{Random multi-control setting}} \\
\midrule
Random    & 0.487 $\pm$ 0.204 & 0.626 $\pm$ 0.212 & 0.821 $\pm$ 0.177 & \textbf{0.757 $\pm$ 0.279}& \textbf{0.766 $\pm$ 0.247}\\
\midrule
\multicolumn{6}{l}{\textit{Text-conditioning strategies under a fixed mask}} \\
\midrule
Rule     & 0.506 $\pm$ 0.188 & 0.648 $\pm$ 0.187 & 0.795 $\pm$ 0.193 & 0.714 $\pm$ 0.257 & 0.715 $\pm$ 0.223 \\
VLM      & \textbf{0.508 $\pm$ 0.196} & \textbf{0.649 $\pm$ 0.197} & \textbf{0.831 $\pm$ 0.174} & 0.728 $\pm$ 0.263& 0.740 $\pm$ 0.231\\
\midrule
GT              & 0.649 $\pm$ 0.152 & 0.776 $\pm$ 0.131 & 0.898 $\pm$ 0.127 & 0.793 $\pm$ 0.196 & 0.828 $\pm$ 0.157 \\
\bottomrule
\end{tabular*}
\end{table*}

After the image-level analysis, we further examine whether the generated results remain reliable after structural parsing. Table~\ref{tab:structure_ablation} summarizes the parser-based structural evaluation. For single structural control signals, more informative conditions generally lead to better parsed results, while simpler controls provide weaker geometric guidance and make structural recovery more difficult.

Under a fixed mask condition, the two text settings still show some difference, which suggests that text guidance continues to affect structural recovery. For the random input setting, the results remain stable when the control condition is randomly selected from five different structural types, indicating that the framework can handle flexible structural inputs without a clear performance drop. We also report the parser performance on real images as a reference upper bound. Overall, CSGen shifts the paradigm of curvilinear synthesis from stochastic approximation to topologically consistent reconstruction, ensuring both high-fidelity details and stability across diverse structural scales.

\subsection{Practical Utility of the Generated Data}

\begin{table}[!t]
\centering
\caption{Downstream gains from synthetic data. Base rows are absolute scores, and augmented rows are changes over base.}
\label{tab:seg_results}
% \small
\setlength{\tabcolsep}{4pt}
\renewcommand{\arraystretch}{1.02}
\begin{tabular}{lcccc}
\toprule
\multirow{2}{*}{Dataset} & \multicolumn{2}{c}{DeepLabV3+} & \multicolumn{2}{c}{SegFormer} \\
\cmidrule(lr){2-3} \cmidrule(lr){4-5}
 & mIoU$\uparrow$ & mDice$\uparrow$ & mIoU$\uparrow$ & mDice$\uparrow$ \\
\midrule
Blood (Base) & 69.62 & 79.37 & 47.89 & 51.00 \\
+ ControlNet & -0.68 & -0.55 & -1.37 & -0.36 \\
+ CSGen & \textbf{+2.64} & \textbf{+2.41} & \textbf{+0.36} & \textbf{+0.68} \\
\midrule
Crack (Base) & 73.86 & 82.68 & 67.87 & 77.58 \\
+ ControlNet & -3.11 & -3.41 & -0.16 & -0.13 \\
+ CSGen & \textbf{+1.71} & \textbf{+2.90} & \textbf{+1.32} & \textbf{+0.69} \\
\midrule
Road (Base) & 79.10 & 87.10 & 61.88 & 70.59 \\
+ ControlNet & +1.21 & +0.72 & +0.56 & +0.66 \\
+ CSGen & \textbf{+4.73} & \textbf{+3.41} & \textbf{+1.16} & \textbf{+1.39} \\
\bottomrule
\end{tabular}
\end{table}

% \begin{figure*}[t]
%   \centering
%   \includegraphics[width=0.8\linewidth]{result.pdf}
%   \caption{Qualitative comparison of segmentation results. From left to right: (a) original image, (b) corresponding mask, (c) segmentation result on the original dataset, (d) segmentation result on the combined dataset with generated samples, (e) segmentation result on generated images, and (f) generated image itself.}
%   \label{fig:data_gen}
% \end{figure*}

To examine the practical usefulness of the generated data, we further conduct a downstream study on three widely used domains,  blood vessels, cracks, and road networks. These datasets are selected because they cover substantially different imaging conditions and appearance statistics, while sharing common structure characteristics such as thin geometry, sparsity, and long-range connectivity. Therefore, they provide a suitable testbed for evaluating whether the generated samples can serve as useful supplementary data beyond the generation task itself. Following this setting, we use two representative segmentation backbones, DeepLabV3+ and SegFormer, and compare the results obtained with and without the generated data. To control for data volume, we also add a same-volume SD3.5+ControlNet synthetic-data baseline, using one generated image per training mask without test-set augmentation or additional filtering. As shown in Table~\ref{tab:seg_results}, ControlNet augmentation is unstable and even decreases performance on Blood and Crack, whereas CSGen consistently improves both backbones across all three domains. These findings indicate that the downstream gains come from the structural quality of the generated samples rather than merely from adding more synthetic data.
% As illustrated in Figure~\ref{fig:data_gen}, the models trained with additional generated data tend to recover finer structural details more clearly, especially in regions with thin or ambiguous curvilinear patterns.
% \FloatBarrier

\section{Conclusion}

In this paper, we propose CSGen, a novel model for the controllable generation of high-quality curvilinear structures. The HPCS resolves the granularity dilemma through a systematic transition from structural anchoring to semantic refinement. Complementing this, the SLRM improves the preservation of fragile, thin branches by dynamically re-weighting the flow-matching loss based on structural sparsity. Crucially, we contribute a high-quality multi-domain dataset, which provides a robust foundation for learning shared structural priors across diverse scenarios. Experiment results across diverse domains show that CSGen achieves superior structure fidelity and semantic adherence. This provides a robust foundation for downstream applications such as data augmentation and reconstruction, bridging a critical gap in curvilinear structure analysis.

\begin{acks}
This work was supported by the National Natural Science of China (No. 62362023, No. 62402354, No. 62406185), the State Key Program of National Natural Science of China (No. 2023YFB4502400), and the Key Research and Development Project of Hainan Province (No. ZDYF2024GXJS313, No. ZDYF2024GXJS262). 

\end{acks}

\bibliographystyle{ACM-Reference-Format}
\balance
\bibliography{ref}

\clearpage
\twocolumn[{
\begin{center}
{\LARGE\bfseries Supplementary Material\par}
\end{center}
}]
\setcounter{section}{0}
\renewcommand{\thesection}{S\arabic{section}}
\renewcommand{\theHsection}{supp.\arabic{section}}
\setcounter{subsection}{0}
\renewcommand{\thesubsection}{\thesection.\arabic{subsection}}
\renewcommand{\theHsubsection}{supp.\arabic{section}.\arabic{subsection}}
\setcounter{subsubsection}{0}
\renewcommand{\thesubsubsection}{\thesubsection.\arabic{subsubsection}}
\renewcommand{\theHsubsubsection}{supp.\arabic{section}.\arabic{subsection}.\arabic{subsubsection}}
\setcounter{figure}{0}
\renewcommand{\thefigure}{S\arabic{figure}}
\renewcommand{\theHfigure}{supp.\arabic{figure}}
\setcounter{table}{0}
\renewcommand{\thetable}{S\arabic{table}}
\renewcommand{\theHtable}{supp.\arabic{table}}

\section{ControlNet}

\begin{figure}[h]
\centering
\includegraphics[width=0.52\columnwidth]{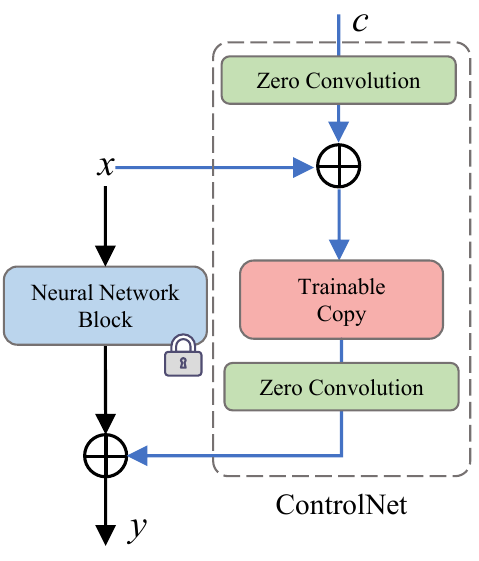} % 
\caption{Illustration of the ControlNet. The original network operates with input $x$ and output $y$. By introducing additional control conditions through an auxiliary structure as shown in the box, the architecture is extended into a ControlNet, enabling controlled generation conditioned on $c$.}
\Description{A schematic diagram of ControlNet showing a neural network block, a trainable copy, zero convolutions, an input control condition, and the controlled output.}
\label{fig:supp_controlnet}
\end{figure}

To enable controllable generation conditioned on layout maps, we adopt the ControlNet framework, as illustrated in Figure~\ref{fig:supp_controlnet}. ControlNet augments a pre-trained neural network by injecting additional control signals $c$ through a parallel branch. Specifically, the original network processes the primary input $x$ to produce the output $y$, while the ControlNet branch applies a pair of zero-initialized convolutions to the control input $c$, ensuring that the control path initially has no influence on the output. The intermediate features from the main network are copied into a trainable sub-network, where they interact with the encoded control signals. The modulated features are then fused back into the main pipeline via additive connections, enabling fine-grained control over the output without altering the backbone. This design preserves the original model’s behavior while supporting conditional generation aligned with $c$.

\section{Dataset}

\subsection{Dataset Analysis}

\subsubsection{Structure Diversity}
The collected domains exhibit substantial variation in thickness, tortuosity, branching pattern, continuity, and sparsity. This diversity is important for training a unified generation model, since it exposes the model to both local boundary variations and global topological changes across domains.

\subsubsection{Public Accessibility}
All datasets are publicly available, ensuring reproducibility and facilitating further research. The consistent availability of open-source data strengthens the generalizability of models trained under this protocol.

\subsection{Control Map Construction}
In the current version of our preprocessing pipeline, we retain four derived control maps: edge, sketch, partial sketch, and width map. All four maps are derived from the standardized binary object mask. The original mask is preserved as the base segmentation control. During training, the structure condition is randomly selected from \{Mask, Width, Sketch, Edge, Partial\}.

\subsubsection{Edge Map}
The edge map is extracted using Canny edge detection on the binary target mask. This representation emphasizes local contour precision and boundary localization, and provides sparse boundary-level structural guidance for controllable generation.

\subsubsection{Sketch Map}
The sketch map is constructed through skeleton-edge structure fusion. Specifically, we first obtain the skeleton of the binary mask as the centerline representation, and then combine it with the edge response to form a unified geometric layout. By integrating centerline and boundary cues, the sketch map provides a more stable structural prior than either component alone.

\subsubsection{Partial Sketch Map}
The partial sketch map is generated from the full sketch map to simulate incomplete structural conditions. Random local regions are removed from the sketch while preserving the main support of the structure, so that the resulting map remains non-empty and structurally valid. This design introduces controlled sparsity and improves the model's robustness to incomplete or occluded structural inputs.

\subsubsection{Width Map}
The width map is generated via distance-based thickness encoding from the binary mask. It explicitly represents the variable caliber of the structures and provides local thickness information that complements the other control conditions.

\subsection{Image Caption}
We explore two complementary strategies: a rule-based captioning method and a vision-language model (VLM)-based approach.

\subsubsection{Rule-Based Descriptors}
We first build a rule-based descriptor from the extracted structural attributes of each sample. Continuous measurements, such as width, tortuosity, branching, and connectivity, are discretized into semantic labels and then organized into a short template-based description with an explicit domain prefix. Compared with free-form text, this descriptor is more stable and consistent across samples, and therefore serves as a reliable structural semantic anchor during training.

\subsubsection{VLM-Based Descriptors}

\begin{table}[t]
\centering
\caption{Prompt structure and content for the VLM-based caption generation pipeline.}
\label{tab:supp_vlm_prompt_structure}
\footnotesize
\setlength{\tabcolsep}{3pt}
\begin{tabular}{p{1.65cm} p{5.6cm}}
\toprule
\textbf{Component} & \textbf{Content} \\
\midrule
System Message &
You are a vision-language annotator for curvilinear structure images. \\
User Prompt &
Generate a detailed environmental description for this curvilinear structure image while remaining consistent with the provided structural prior. \\
Prior System Prompt &
Inject quantified metadata and structural attributes as prior constraints to reduce hallucination and preserve consistency with the underlying structural truth. \\
Domain Prefix &
Use an explicit domain prefix in the prompt to distinguish different curvilinear domains. \\
Appearance Instructions &
Describe appearance, background texture, contrast, foreground intensity, background intensity, texture complexity, and imaging style. \\
Constraint &
Do not introduce hallucinated or structurally inconsistent content. The generated description should remain synchronized with the quantified metadata and the target structure. \\
Language Constraint &
Generate a coherent natural description suitable for diffusion-model training. \\
\bottomrule
\end{tabular}
\end{table}

To complement the rule-based descriptors, we further employ a VLM to generate detailed environmental descriptions. Unlike the rule-based branch, which mainly captures stable geometric semantics, the VLM-based branch focuses on fine-grained appearance and context information. To prevent the VLM from generating hallucinated or structurally inconsistent text, we inject the precisely quantified metadata into the VLM as a prior system prompt, as shown in Table~\ref{tab:supp_vlm_prompt_structure}. This constraint-aware prompting ensures that the generated captions remain synchronized with the underlying structural truth. The resulting descriptions are then used as the final T5 prompts in the hierarchical text-conditioning pipeline.

\section{Experiment}

\subsection{Hyperparameter Settings and Environment}

\begin{table}[t]
\centering
\caption{Key hyperparameter settings for generation and segmentation experiments.}
\label{tab:supp_exp_settings}
\small
\setlength{\tabcolsep}{8pt}
\begin{tabular}{lcc}
\toprule
\textbf{Setting} & \textbf{Generative} & \textbf{Segmentation} \\
\midrule
Optimizer & AdamW & SGD \\
Initial LR & $5 \times 10^{-6}$ & 0.001 \\
LR schedule & Constant + warmup & Poly \\
Warmup steps & 500 & -- \\
Weight decay & $1 \times 10^{-2}$ & $5 \times 10^{-4}$ \\
Batch size & 1 & 4 \\
Accumulation & 1 & -- \\
Epochs & 48 & 150 \\
T5 schedule & start at 0.4, full at 0.8 & -- \\
Loss weighting & SLRM & -- \\
Max sequence length & 512 & -- \\
Precision & bf16 & fp32 \\
Checkpoint freq. & 500 steps & 1 epoch \\
Validation freq. & 500 steps & 1 epoch \\
\bottomrule
\end{tabular}
\end{table}

All experiments are run on a Linux server equipped with 8 NVIDIA L20 GPUs (48 GB each), an Intel Xeon Platinum 8260 CPU (48 cores), and 512 GB RAM. To ensure full reproducibility, we fix the global seed to 2026 at the start of each run. The table provides an overview of the core hyperparameter configurations, which are applied uniformly across all experiments to guarantee reproducibility.

\subsubsection{Generation Model}
Text conditioning uses two CLIP encoders and one T5 encoder. The CLIP encoders produce token-level hidden states and global projections that are fused into a joint embedding. The T5 encoder refines the token sequence into contextualized features. All outputs are aligned and concatenated for cross-attention and timestep embedding.

Noise scheduling employs a flow-matched Euler discrete scheduler to discretize timesteps and define variance updates, ensuring stable diffusion.

AutoencoderKL maps images to latent feature maps and reconstructs RGB outputs; its parameters remain fixed during fine-tuning.

Stable Diffusion 3.5 adopts a multimodal diffusion transformer (MM-DiT) architecture. Unlike previous U-Net-based backbones, MM-DiT employs separate sets of weights for image and text modalities, allowing more complex cross-modal interaction through joint attention blocks. In our framework, the frozen SD3.5 backbone provides pretrained generative knowledge, while ControlNet incorporates external conditioning signals from structural maps. During training, only the ControlNet branch is optimized with the flow matching objective. In addition, we adopt the sparsity-aware loss re-weighting mechanism (SLRM) to improve the optimization focus on thin and fragile structures.

\subsubsection{Segmentation Model}
For parser-based structure evaluation, we employ SegFormer as the segmentation parser. Input images are normalized and resized to a fixed resolution before being fed into the network. SegFormer adopts a hierarchical vision-transformer encoder that extracts multi-scale features through overlapping patch embeddings and successive transformer blocks. The decoder then aggregates features from different stages to produce the final segmentation map. During inference, the predicted masks are compared with the corresponding reference masks to compute mIoU, mDice, tprec, tsens, and clDice.

For the downstream utility study, we further use DeepLabV3+ and SegFormer as representative segmentation models. DeepLabV3+ employs a ResNet-50 backbone with dilated convolutions and an atrous spatial pyramid pooling module to capture multi-scale context, followed by a decoder that fuses high-level and low-level features for dense prediction. SegFormer serves as a transformer-based counterpart with a lightweight multi-scale decoding design. These two models are used to evaluate whether the generated samples can provide consistent improvements for downstream curvilinear structure segmentation.

\end{document}